# A Visual Entity-Relationship Model for Constraint-Based University Timetabling

Islam Abdelraouf, Slim Abdennadher, Carmen Gervet

Department of Computer Science, German University in Cairo
[islam.abdelraouf, slim.abdennadher, carmen.gervet]@guc.edu.eg
`http://met.guc.edu.eg`

**Abstract.** University timetabling (UTT) is a complex problem due to its combinatorial nature but also the type of constraints involved. The holy grail of (constraint) programming: "the user states the problem the program solves it" remains a challenge since solution quality is tightly coupled with deriving "effective models", best handled by technology experts. In this paper, focusing on the field of university timetabling, we introduce a visual graphic communication tool that lets the user specify her problem in an abstract manner, using a visual entity-relationship model. The entities are nodes of mainly two types: resource nodes (lecturers, assistants, student groups) and events nodes (lectures, lab sessions, tutorials). The links between the nodes signify a desired relationship between them. The visual modeling abstraction focuses on the nature of the entities and their relationships and abstracts from an actual constraint model.

## 1 Introduction

University timetabling (UTT) is a complex problem due to its combinatorial nature but also the type of constraints involved. Several approaches have been proposed to solve timetabling problems for specific instances using several approaches, e.g. [5,3,1].

One of the shortcomings of the current constraint-based systems is the existence of modeling tools/languages to express constraint problems. Several approaches have been proposed to overcome these problems. A lot of work has been invested to propose languages that allow the specification of a combinatorial problem at a high level of abstraction, e.g. [2]. Alternatively, visual modeling languages have been proposed to generate constraint programs from visual drawings, e.g. [4]. In general, the visual drawings correspond to constraint graphs where the nodes describe the variables of the problem with their associated domains and the edges correspond to the constraints between each pair of variables. The constraint graphs provide a visual counterpart to constraint satisfaction problems. However, in practice they are intractable for real world problems even if abstraction is introduced into the constraint graph.

In this work, we propose an orthogonal approach where the user models the constraint problem visually by drawing a graph that defines the available resources and the tasks to be scheduled. The graph does not describe the constraints explicitly. It consists of nodes and links, where the nodes are mainly of two types: resource nodes (lecturers, assistants, student groups) and events nodes (lectures, lab sessions, tutorials). The links describe relationships between the nodes. Depending on the type of node, the semantics of the links is determined.

This approach enjoys three main properties: 1) the problem is stated at a high level in a constraint and variable free manner, 2) online preliminary consistency checks of proposed links are possible thanks the rich semantic carried by the nodes, 3) a compilation into an effective constraint programming model including global constraints is performed using the properties of the nodes and specified links.

Our system was built in Java and compiled into SICStus Prolog. Tests were run to build a complete timetable for the German University in Cairo (GUC) including over 200 events to be scheduled and over 400 resources in 3 minutes.

This paper is organized as follows. In Section 2, we present the GUC timetabling problem. In Section 3, we describe how the problem can be modeled as a constraint satisfaction problem.

Section 4 presents the visual graphic communication tool for generic timetabling problems. In Section 5, we address additional problem components that are particular for the GUC. In Section 6, we discuss the scalability and the modularity of the approach. Finally, we conclude with a summary and directions for future work.

## 2 GUC Timetabling

The GUC consists of four faculties. Each faculty offers a set of majors. Currently at the GUC, there are 140 courses offered and 6500 students registered for which course timetables should be generated every semester. There are 200 staff members available for teaching. Each faculty offers a set of majors. For every major, there is a set of associated courses. Faculties do not have separate buildings, therefore all courses from all faculties should be scheduled taking into consideration shared room resources.

Students registered to a major are distributed among groups for lectures (lecture groups) and groups for tutorials or labs (study groups). A study group consists of maximum 25 students. In each semester, study groups are assigned to courses according to their corresponding curricula and semester. Due to the large number of students in a faculty and lecture hall capacities, all study groups cannot attend the same lecture at the same time. Therefore, sets of study groups are usually assigned to more than one lecture group. For example, if there are 27 groups studying Mathematics, then 3 lecture groups will be formed.

The timetable at the GUC spans a week starting from Saturday to Thursday. A day is divided into five time slots, where a time slot corresponds to 90 minutes. An event can take place in a time slot. This can be either a lecture, tutorial, or lab session and it is given by either a lecturer or a teaching assistant. Lectures are given by lecturers and tutorial and lab sessions are given by teaching assistants (TA). In normal cases, lectures take place in lecture halls, tutorials in exercise rooms and lab sessions take place in specialized laboratories depending on the requirements of a course. In summary, an event is given by a lecturer or a teaching assistant during a time slot in a day to a specific group using a specific room resource. This relationship is represented by a timetable for all events provided that hard constraints are not violated. These constraints cannot be violated and are considered to be of great necessity to the university operation. The timetable also tries to satisfy other constraints which are not very important or critical. Such constraints are known as soft constraints that should be satisfied but may be violated. For example, these constraints can come in form of wishes from various academic staff.

Some courses require specialized laboratories or rooms for their tutorials and lab sessions. For example, for some language courses a special laboratory with audio and video equipment is required. The availability of required room resources must be taken into consideration while scheduling. Some lecturers have specific requirements on session precedences. For example, in a computer science introductory course a lecturer might want to schedule tutorials before lab sessions.

Furthermore, some constraints should be taken into consideration to improve the quality of education. One of the constraints requires that a certain day should have an activity slot for students, and a slot where all university academics can meet. For those slots no sessions should be scheduled. A study group should avoid spending a full day at the university. In other words, the maximum number of slots that a group can attend per day is 4. Therefore, if a group starts its day on the first slot, then it should end its day at most on the fourth slot. Furthermore, though the university runs for 6 days a week each study group must have at least two days off which means that a study group can only have sessions for five days a week.

A certain number of academics are assigned to a course at the beginning of a semester. Teaching assistants are assigned to one course at a time. For courses involving lab and tutorial sessions, a teaching assistant can be assigned to both or just one of them. This should be taken into consideration when scheduling to avoid a possible clash. Furthermore, the total numbers of TAs assigned to a session should not exceed the maximum number of assistants assigned to the corresponding course at any time. A lecturer can be assigned to more than one course. This should be considered when scheduling in order to avoid a possible overlap. Academics assigned to courses have a total number of working and research hours per day and days off that need to be considered when scheduling. Courses should be scheduled in a manner such that the number of session hours given by an academic should not exceed his or her teaching load taking into consideration days off.

## 3 Modeling UTT as a Constraint Satisfaction Problem

A Constraint Satisfaction Problem (CSP) is a tuple $(X, D, C)$ where $X$ is a set of variables $\{x_1, ..., x_n\}$, $D = \{D_1, ..., D_n\}$ a set of associated domains for all $x_i \in X$, and $C$ a set of constraints [6].
In the CSP model of the UTT, we define the variables $\mathbf{E}$ (possibly subscripted) to be lecture slots, tutorial slots, and lab slots taken by groups. Since we have 5 slots in a day and 6 days in a week, the domain is [0..29]. Slot 0 would correspond to the first slot on Saturday (first day of the week in Egypt), and slot 22 would correspond to the third slot on Wednesday. Consequently, slots from 0 to 4 correspond to Saturday, 5 to 9 correspond to Sunday, etc. The goal is to find an allocation of one slot per variable such that the problem resource and scheduling constraints hold.
The generic UTT constraints are of three kinds: 1) temporal constraints requiring precedences among events ($\mathbf{E_i} \leq \mathbf{E_j}$), 2) resource constraints requiring events sharing some resources not to overlap in time, 3) and university specific constraints such as preferences for lecturers and teaching Assistants.
Note that resource constraints can take two forms in the timetabling problem:

1. when different events share a single resource, like a number of lectures taught by one lecturer, in this case none of those events can overlap. They can be modeled using inequality ($\mathbf{E_i} \neq \mathbf{E_j}$) among events associated with the same resource or the `all_different`($[\mathbf{E_i}, \mathbf{E_j}]$) global constraint,
2. when a set of events share multiple resources. This generalizes the previous form and it occurs in the UTT problem for instance when a set of events require the same type of rooms (halls or labs) and there is a limited amount of them. A global constraint can be used to enforce this which is the `cumulative/2` constraint. In SICStus Prolog, the `cumulative` constraint takes a list of `task` predicates defined by their start time, duration, end time, resource usage and identifier. Its second argument is a list containing the amount of resource available. It will be used in our code generation.

The last type of constraints applied are university specific constraints which for the GUC relate to the extra day off (not bound to Saturday) and the maximum allowed slots per day for study groups. A day off would mean a day with no events scheduled for that lecturer or TA. This can be expressed as follows if we consider Saturday as the day-off for a given lecturer. All the lectures relative to a lecturer(we denote by $L_i$ the list of events associated with lecturer $i$) must start after time slot 4:

$$\forall_j \mathbf{E_j} \in L_i \Rightarrow \mathbf{E_j} > 4$$

In general, timetabling is an over-constrained problem. Thus, the main aim is to find the best solution that satisfies all hard constraints and as many soft constraints as possible. For the German University in Cairo, different lecturers have different requirements in terms of days off and time slots they would like to have free. Thus, the best solution for the German University In Cairo is the one that satisfies the largest number of the lecturers' wishes. These soft constraints are modeled by associating a flag with each wish that determines whether the wish is satisfied or not. For example, the constraint that a lecturer $i$ would like to have the first slot on Saturday free can be modeled as follows:

$$0 \notin L_i \Longrightarrow P_k = 1$$

where $P_k$ is the flag associated with the $k^{th}$ wish.

The best solution for the GUC would be the one that maximizes the number of satisfied wishes, i.e., the solution maximizes a cost function that is defined as follows:

$$SCORE = \sum_{n=1}^{n=m} P_n$$

where m is total number of wishes .

## 4 From a visual graph to a CSP program

Our goal is to offer the end-user a tool that would allow her to specify the problem as she sees it, while enabling us to derive from it an effective constraint model. To do so we chose the concept of graph as a natural vehicle of connections or relationships between multiple entities or nodes. In this section we define the components that define a graphical specification elaborating on the semantic of the nodes which is carried visually and internally as a typed data structure. It applies to the generic components of UTT problems. We then give the modular structure of the tool and show how a graphical specification is compiled into a SICStus program. We finally develop the specification of additional problem components that are particular to a given university and are specified using other elements of our tool.

### 4.1 Visual graph components

The graph is not a constraint graph in the sense that nodes are not variables and edges constraints. The visual structure of a graph has been chosen to ease the abstract specification of the problem. Its visual nature helps convey the problem structure to third party (faculty deans, administration,etc). It is structured in a way that aids in both problem specification and future compilation into a CP model. Elements of the timetabling problem are available resources (lecturers, teaching assistants, study groups), courses, and different course events. The graph is composed of nodes and undirected links. As the user constructs her visual graph to specify her problem, certain information must be held at the node to aid in the dynamic CP model generation. In particular the nodes specify different components of a UTT. Let us first describe the different types the nodes can denote.

*Nodes.* Nodes can be of three types:

1. A resource. These are lecturers, study groups or teaching assistants.
2. An event. These are lectures, tutorials or lab sessions.
3. Meta-events. These are the courses. A course is often composed of a lecture and a tutorial and possibly some lab sessions. Thus, the end-user can define a course and associate with it the relevant events it involves. This type of node does not appear in the generated CSP model, it is used to ease the visual specification of the problem.

Each node instance corresponds to a specific icon, carrying out its semantics from the end-user viewpoint. To illustrate drawing facilities of the system, we give a snapshot of a small graph being created. The end-user defined two courses (meta-events), Math and Physics, connected to 3 and 1 events, respectively. The Math course comes with 2 tutorial sessions and one lecture, whereas the Physics course comes so far with only one lecture. In Fig. 1, we present the system with the drawing graph components. On the left hand side, the node icons are available. On the right hand side, the drawing pad, where the graph is created, is available. Below the node icons, the node properties to be added once a node has been inserted into the graph tab is available.
The icon indicates the type of a node and properties associated with it. For instance, lectures and tutorials require certain room size; lectures require lecture halls while tutorials require classrooms. Representing this on the diagram would require creating an icon or a node for each room type and connecting this node to all events that require this type of node. Since all events require a certain type of room, displaying the rooms as an extra icon would increase the number of edges and make the graph harder to trace. Instead of connecting events that share a certain type of rooms to a single resource node, all events requiring the same room resource have similar icons (lectures have the same icon which is different from that of tutorials and labs). Additional properties are filled in for each created node in the property tab.

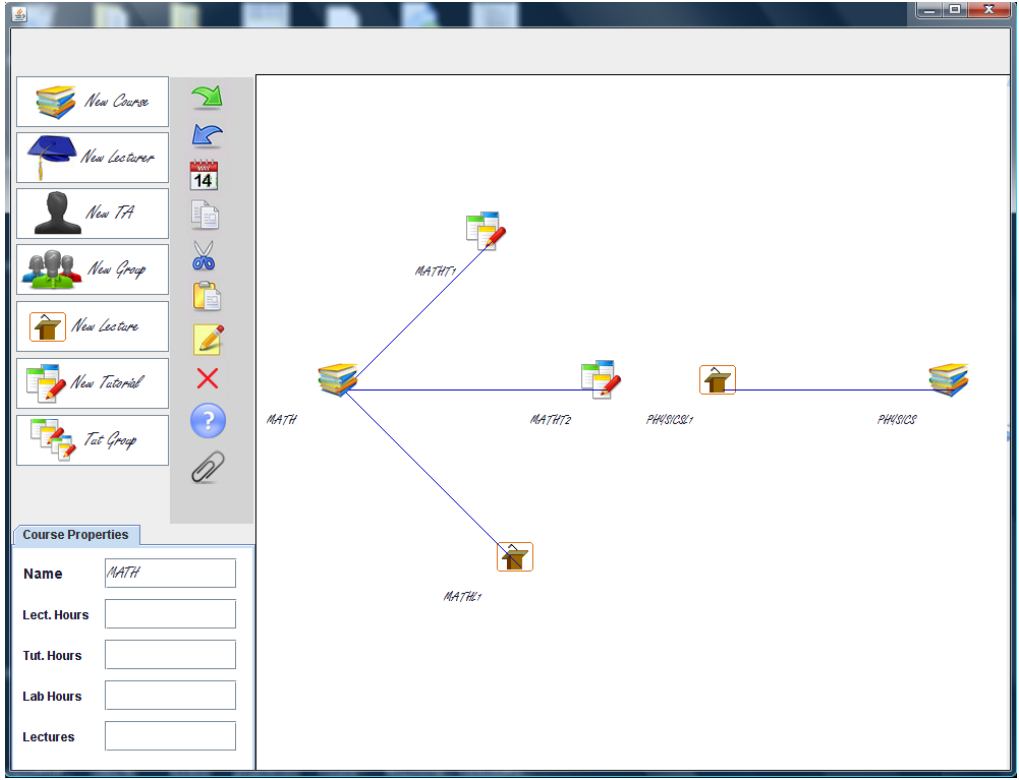

**Fig. 1.** Created graph with two courses and 4 events

*Links.* Links between nodes are undirected connections, signifying a relationship between two nodes (ie. a resource will carry out an event, or a course is defined by one lecture and 3 tutorials). The approach is user-friendly, but could be prone to specification errors. Thus while the end-user does not deal with node types explicitly, the system performs a dynamic checking at link creation to ensure that resource-resource links can not be validated, and resource-event links are only created if the event suits the type of the resource (e.g. a lecturer can only teach lectures no tutorials or labs). This is described below.

### 4.2 Graph semantic check

During the creation of the graph, we perform a dynamic semantic check that prevents incorrect link requests from being drawn between two nodes. While the user can add any number of nodes, certain connections would not make sense.
The semantic check is performed as the end-user attempts to connect two previously created nodes. When the user attempts such a connection, a request is sent that ensures that:

– A course can only be connected to lectures, tutorials or labs nodes.
– A lecture node can only be connected to one lecturer, one course and study groups.
– A tutorial or lab node can only be connected to one TA, one course and one group.
– A group node can only be connected to event nodes (lectures, tutorials or labs).
– A lecturer node can only be connected to lectures nodes.

– A TA node can only be connected to tutorials or labs nodes.

Any other connection request is ignored and no line can be drawn between the desired nodes.

*Node implementation.* At the implementation level, each resource node is associated with a dynamic list of events this resource node is currently connected to. For instance a lecturer node would be linked to a list of lectures, the person is assigned to. Each event node (Lecture, tutorial, lab) is linked with two flags: one flag stating whether this node is connected to its corresponding event (lecture to a lecturer, lab to a TA, etc). This flag is primarily used to prevent a single event node from being connected to multiple resource node of the suitable type. The second flag states whether the event node is connected to a course and is used to prevent a single event from being connected to multiple courses. A static information is also attached to each resource node which is the constraint predicate the list will be applied to.

### 4.3 Generic CP code generation

This is a core component of the system. The main objective is to generate an efficient CP code from the visual graph and node properties that goes beyond binary constraints and includes global constraints. Our code generation module is based on the following observations: 1) links involving resource nodes drive the model. Thus when a resource node is created in the graph we create a dynamic list in the CSP model that will contain all the events connected to it, 2) all the established connections are semantically viable because the dynamic checks at link creation have been carried out beforehand.

Since the constraints are derived from the relations between nodes then a node must be aware of all its direct neighbors. In other words, each node contains a list of nodes representing its direct neighbors. Each node has a type from which we know the type of constraints to be applied.

*The alldifferent constraint* . Consider the graph of Fig. 1, there is no resource node involved at this stage, thus no constraints are generated yet. Adding the lecturer as illustrated in Fig 2. will constrain events that share a common resource and trigger the code generation.

This establishes a resource dependency link between event nodes. From a constraint point of view, this implies that the two lecture events (MATH and PHYSICS) share a common resource, thus can not occur simultaneously. The system generates a list of event variables (corresponding to the events connected to a resource node) and constrains all the variables in this list to be pairwise distinct. The most efficient constraint available to this date in CP systems to enforce this restriction is the `all_different/1` constraint. It takes as input a list of event variables that cannot share a value. In our case, the domain of event variables is the time slot at which it can occur. Note that the initial domains are defined as the system is initialized, since the user will enter the number of slots per day and the number of days per week. In the case of the GUC we have $5 \times 6 = 30$ (with domains [0..29]). The following code block will be generated as the lecturer node is connected:

```
LECTURER1 = [MATHL1, PHYSICSL1],
all_different(LECTURER1),
```

Now if the user adds resource nodes relative to the study groups as illustrated in Fig. 3, two blocks of code will be generated relative to these resources.

In this case, the code block generated on creation of the node `GROUP1` and its relative links, is:

```
GROUP1 = [MATHL1, MATHT1, MATHT2],
all_different(GROUP1),
```

and the code generated by `GROUP2` will be:

```
GROUP2 = [PHYSICSL1],
all_different(GROUP2),
```

Note that the code is updated dynamically, which means that as a new event is connected to a given resource node, the list of events associated to this node is updated and so is the relative code block.

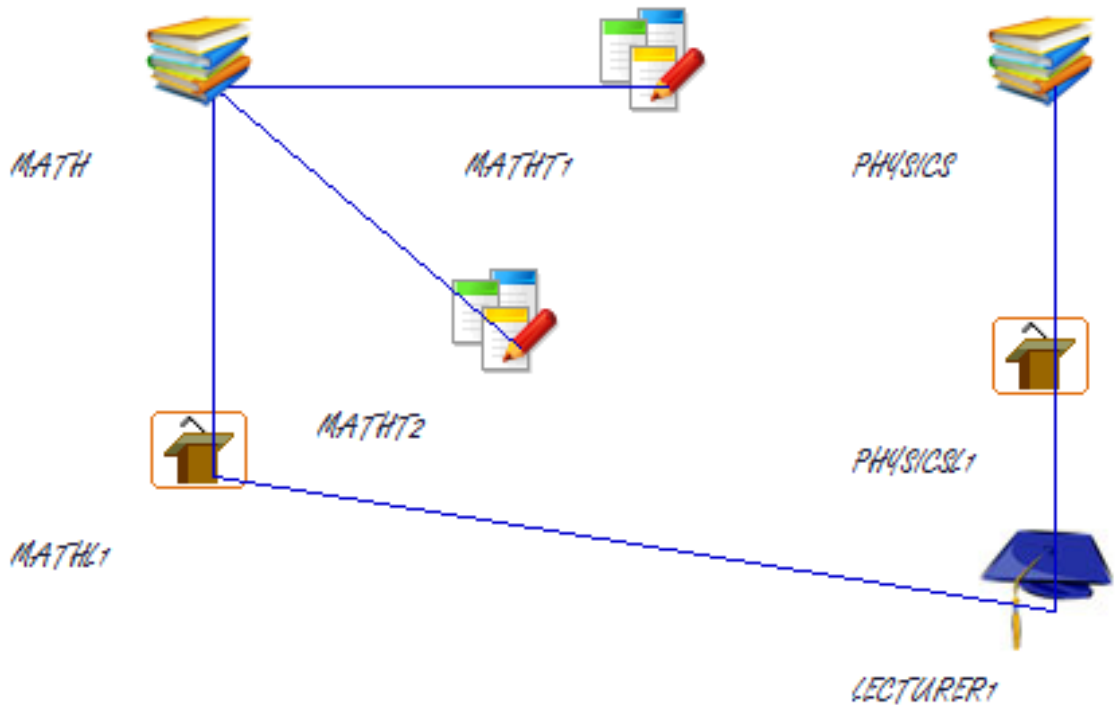

**Fig. 2.** Two courses talked by one lecturer

*The cumulative constraint* . Lectures and tutorials require specific rooms: lectures require lecture halls while tutorials require classrooms. As mentioned before, all events requiring the same room type (or resource) have similar icons. The most effective way to constrain events that share a finite number of resources is the global cumulative/2 constraint. This constraint takes as input a set of event variables with possible start dates as domain, and number of possible resources to share. It ensures that the number of overlapping events requiring a common room type should not exceed the number of rooms available of that type. We generate the code block using the cumulative constraint for each room type existing in the university and each event is included in its corresponding cumulative constraint according to its type.

The program generation is done by concatenating blocks of code generated by node with the cumulative constraints and the domain constraint. The block code generated is as follows:

```
domain([MATHT1E0, MATHT2E1], 1, 4),
%% each event last one slot and consumes 2 rooms,
%% there are atmost 2 rooms available in this example
cumulative([task(MATHT1, 1, MATHT1E0, 2, 0),
task(MATHT2, 1, MATHT2E1,  2, 1)],
[limit(2), global(true)]),

domain([PHYSICSL1E0, MATHL1E1], 1, 4),
cumulative([task(PHYSICSL1, 1, PHYSICSL1E0, 2, 0),
task(MATHL1, 1, MATHL1E1, 2, 1)],
[limit(2), global(true)]),
```

The system with the draw tab allows the user to visualize the generated code as illustrated in Fig 4.

The window is divided into three major sections; the first contains the graph, below it are two other windows, one appears whenever the user clicks on a node which allows the user to provide simple information that is node specific like the node name. The third window shows the user the generated SICStus program. The model specified and generated so far is generic.
However, a UTT problem deals with university specific constraints and preferences and these should be part of the visualization and the CP model. We present two other features of our system, namely the time tab, and the preference tab.

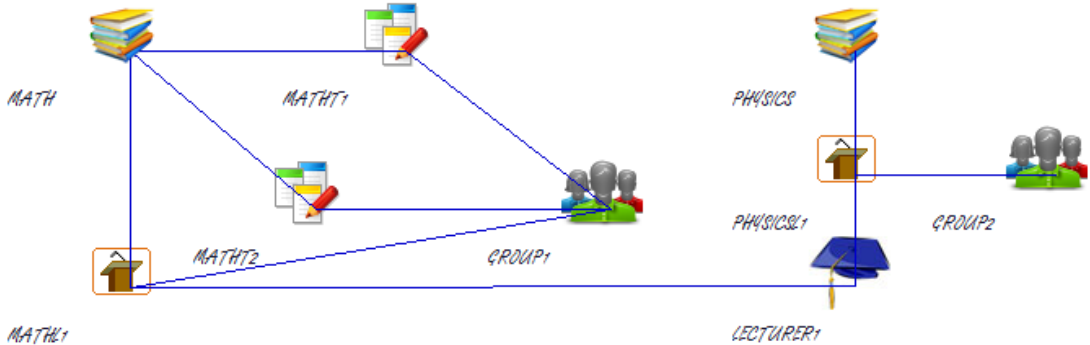

**Fig. 3.** Addition of study groups

## 5   University specific constraints

The use of specific icons indicates the type of room an event node is related to. This eliminated the need for a room resource node. However, to ensure that there is enough room resources available, we must know the capacity of each room resource, and we must know the number of rooms of that type. These informations can not be acquired from the graph. Therefore we provide complementary means for the user to provide additional information. The user can choose between different tabs. The first tab described previously deals with the graph construction. We now present the preference tab and the time tab.

### 5.1   Dealing with preferences

The second tab (Fig. 5) enables the user to tailor the problem to a specific university and set preferences (room availability, extra day-off considered or not, full day for student groups or not). All node types can access the preference tab and thus information can be exchanged. If the day off for a study group is selected a corresponding code block will be generated. The following block of code will be generated as for LECTURER1. It is a hard constraint then ensures that only one extra day-off is allowed per resource node.
The `count_interval` is a user defined constraint that constrains the number of events per resource (list LECTURER1) that fall in the time interval (eg. [0,2]) to be LECTURER1D0C.

```
count_interval(0, 2, LECTURER1, LECTURER1D0C),
LECTURER1D0C #\= 0 #<=> LECTURER1D0,
count_interval(3, 5, LECTURER1, LECTURER1D1C),
LECTURER1D1C #\= 0 #<=> LECTURER1D1,
count_interval(6, 8, LECTURER1, LECTURER1D2C),
LECTURER1D2C #\= 0 #<=> LECTURER1D2,
LECTURER1D2 + LECTURER1D1 + LECTURER1D0 #< 3,
```

Also, according to the GUC policy, a student should not have a full day which means a student having a class on the first time slot of a day can not have a class on the last time slot of the same day. Ensuring this would mean that on each day a study group can either have a class on the first time slot, or a class on the last time slot of that day but not both. Applying this to our problem would result on the following constraints being generated as the user creates the study group nodes. The `count` predicate is a SICStus constraint. If the first line below it can be read as: the list of events in `GROUP1` is constrained to have `GROUP1S0C` variables (Boolean) with value "0" (meaning taking place in the first slot of Saturday). It also constrains the events to take place only in the first or last slot of the day (line 3):

```
%% Group1 will generate
count(0, GROUP1, GROUP1S0C),
count(2, GROUP1, GROUP1S2C),
GROUP1S0C + GROUP1S2C#=<1,
count(3, GROUP1, GROUP1S3C),
count(5, GROUP1, GROUP1S5C),
GROUP1S3C + GROUP1S5C#=<1,
count(6, GROUP1, GROUP1S6C),
count(8, GROUP1, GROUP1S8C),
GROUP1S6C + GROUP1S8C#=<1,

%% The same code will be generated for all groups
```

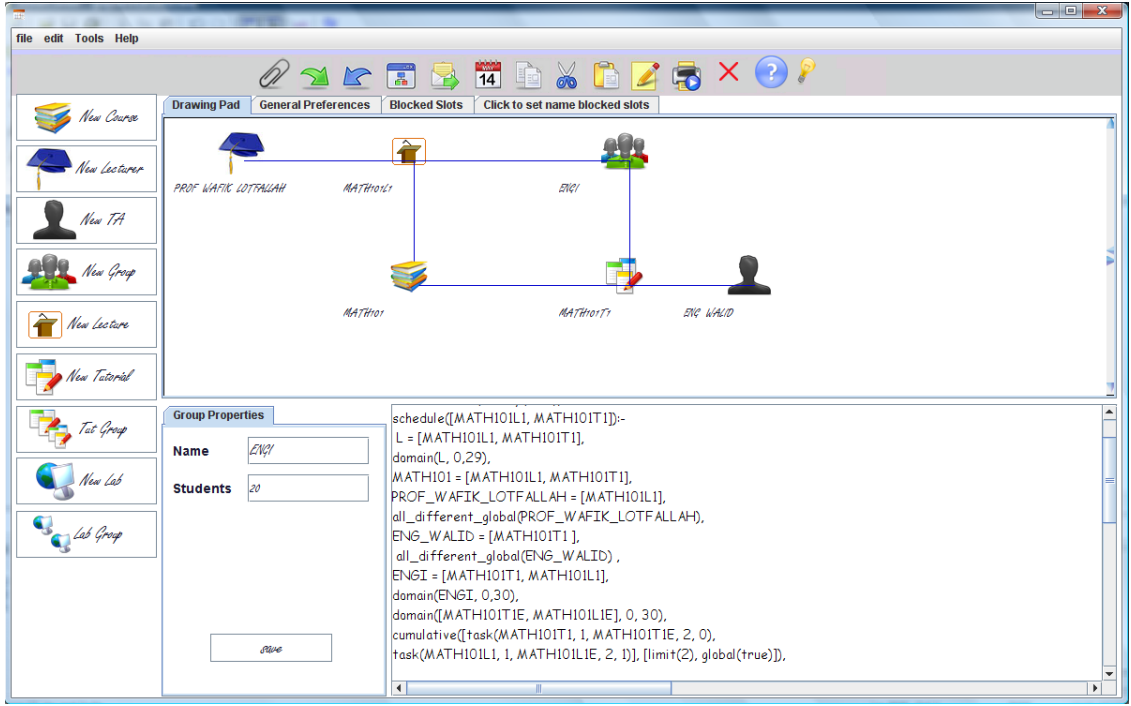

**Fig. 4.** The system with the draw tab

### 5.2 Dealing with time

The German University in Cairo requires lecturers to have Friday off and another day off (not necessarily Saturday). Thus, each lecturer should have one extra day where there is no scheduled events. Applying this to our problem would mean that either no lectures are given on the first day or no lectures are given on the second day or no lectures are given on the third day, etc. The end-user specifies the lecturer's day off using the following tab in Fig. 6. It shows the time line and allows the user to block slots by simply clicking on the required slot.

If a lecturer wishes that no lectures should be given in the first slot on Saturday as illustrated in Fig. 6, then the following fragment of code will be generated:

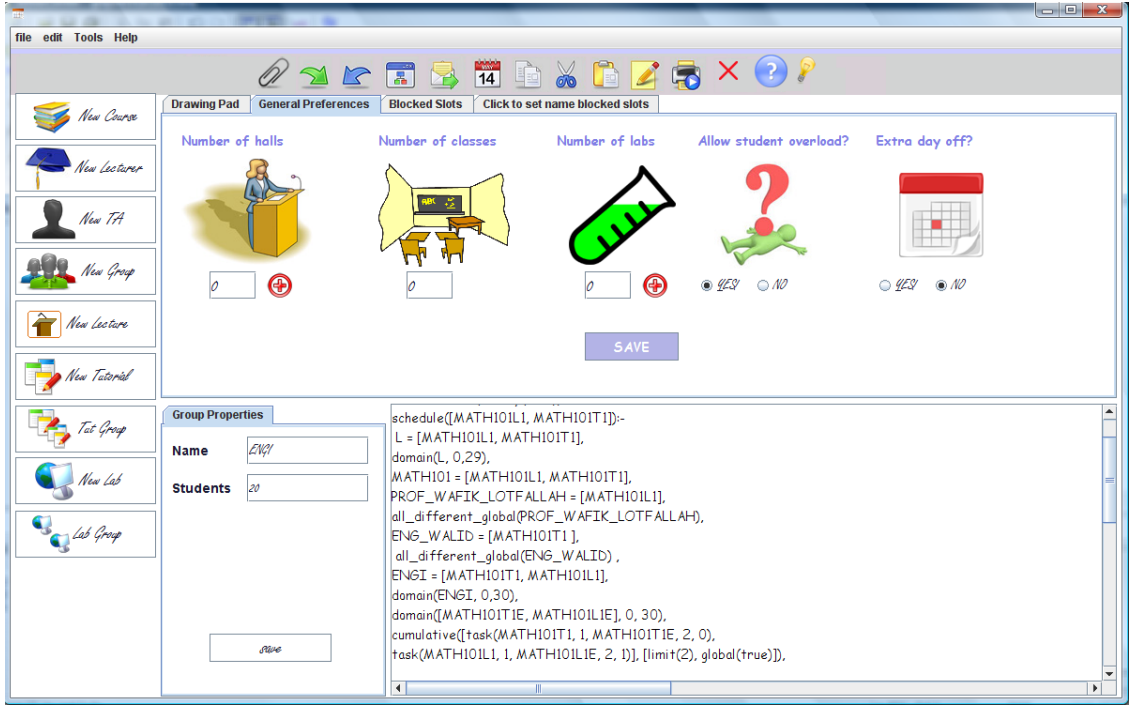

**Fig. 5.** The system with the preferences tab selected

```
count(0, DR_WAFIK_LOTFALLAH, SCOUNT0),
SCOUNT0 #= 0 #<=> SCON0,
SCONS #= SCON0,
```

SCOUNT0 is the number of 0s in the list of the lecturer DR_WAFIK LOTFALLAH. SCOUNT0 being equal to 0 would result in the flag $SCON0$ being set to 1. SCONS is supposed to be the sum of the satisfied soft constraints. In this example, we only have one soft constraint thus $SCONS$ is equal to SCON0. As mentioned in Section 3, the aim of the timetabling is to find a solution that maximizes the number of lecturers' wishes. Thus, the labeling part will be modified as follows:

```
labeling([ffc, maximize(SCONS)],L).
```

Where `fcc` is the heuristic to select the variable with the smallest domain, breaking ties by selecting the variable that has the most constraints suspended on it and selecting the leftmost one.

## 6  System scalability and evaluation

In addition to the use of global constraints, a key asset of this approach is its modularity and scalability. With a focus on the resources and their direct links we can easily change properties of any created icon or add new links to the graph. This will lead to a dynamic update of the CP model. From the constructed graph, direct and indirect relations (e.g which lecturer teaches which study group) between events and resources can be clearly understood. Looking at the diagram one can clearly see all the components of a UTT instance. This is not the case if the problem is modelled as a constraint network [4]. Fig. 7 represents a constraint network with four variables. From the network, one can clearly understand that variables COURSE1L1 and COURSE2L1 can not be equal. However, one cannot understand the reason behind it, since on the network there is no difference between these two variables and any other two unequal variables in their representation. Either the problem description or the constructed problem model should be consulted to understand why such relation exists. Since the people involved in the university timetabling process are usually not programming experts, a visual representation would require a clear representation of the problem through which the problem can be clearly understood by all stakeholders which is the case using our visual entity-relationship model.

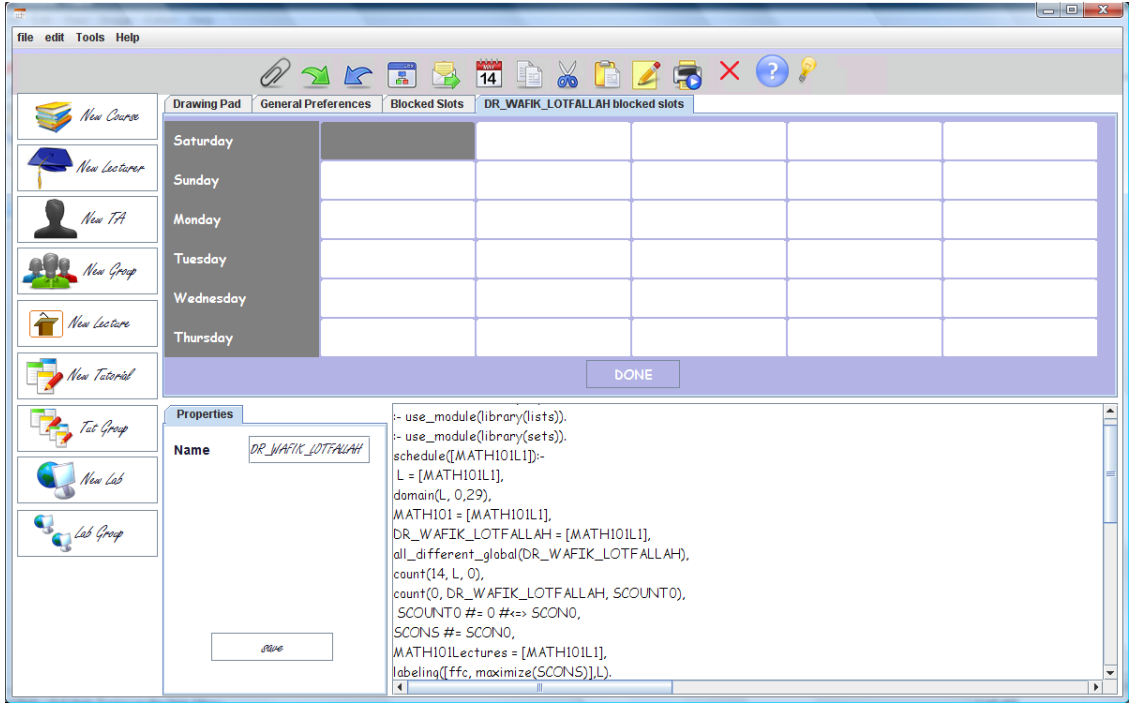

**Fig. 6.** The time tab of a lecturer with a single slot blocked

The system was used to generate the GUC university timetabling for one term. It was run on an Intel(R) Core(Tm) 2 Duo , with CPU 2.4 GHz with 2GB ram. The problem corresponds to 2233 events to be scheduled, over 400 resources, including 224 study groups and 201 lecturers and teaching assistants.

For this instance, the visual entity-relationship model was drawn in 5 hours. The corresponding SICStus code was generated in 7 seconds from the constructed graph. The first 5 solutions were found after three and half minutes. However, a manually generated schedule takes in general two to three months to be constructed by one timetable specialist.

## 7   Conclusion and Future Work

We have presented a visual graphical approach to specify combinatorial problems with application to university timetabling. The main contribution lies in the usage of type nodes to be distinguished from constraint graph variable nodes, and represent resources and events of the problem at hand. The rich semantic of the nodes and the links created among them can be used to generate efficient CP models using global constraints in a dynamic setting.

We intend to improve the representation of the graphs by providing additional features to group a number of icons and to combine them into one icon with the ability to expand and collapse that icon. Additionally, an interesting direction for future work is to investigate how the proposed approach can be generalized to handle any resource allocation problem by adding a generic interface to express application specific constraints.

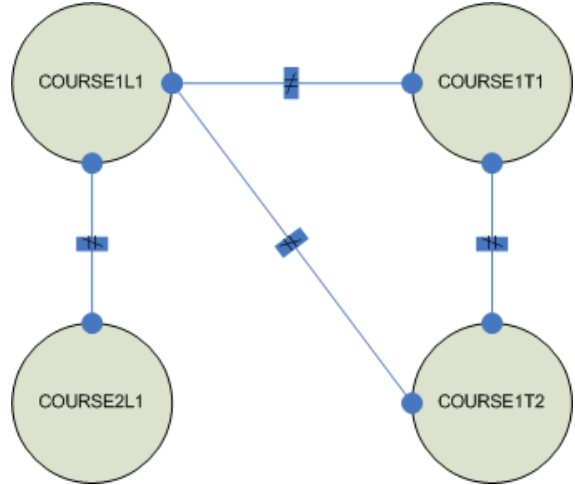

**Fig. 7.** A constraint network representing the timetabling problem